\documentclass[conference]{IEEEtran}
\IEEEoverridecommandlockouts
\usepackage{cite}
\usepackage{amsmath,amssymb,amsfonts}
\usepackage{algorithmic}
\usepackage{graphicx}
\usepackage{textcomp}
\usepackage{xcolor}

\usepackage{array}
\usepackage{multirow}
\usepackage{geometry}
\usepackage{longtable}
\usepackage{url}
\usepackage{booktabs}

\def\BibTeX{{\rm B\kern-.05em{\sc i\kern-.025em b}\kern-.08em
    T\kern-.1667em\lower.7ex\hbox{E}\kern-.125emX}}
\begin{document}

\title{Addressing Small and Imbalanced Medical Image Datasets Using Generative Models: A Comparative Study of DDPM and PGGANs with Random and Greedy K Sampling}

\author{\IEEEauthorblockN{Iman Khazrak\IEEEauthorrefmark{1},
Shakhnoza Takhirova\IEEEauthorrefmark{2}, Mostafa M. Rezaee\IEEEauthorrefmark{3}, Mehrdad Yadollahi\IEEEauthorrefmark{4}, \\ Robert C. Green II\IEEEauthorrefmark{5},
Shuteng Niu\IEEEauthorrefmark{6}}
\IEEEauthorblockA{Department of Computer Science,
Bowling Green State University\\
Bowling Green, OH, USA\\
Email: \IEEEauthorrefmark{1}ikhazra@bgsu.edu,
\IEEEauthorrefmark{2}takhirs@bgsu.edu,
\IEEEauthorrefmark{3}mostam@bgsu.edu,
\IEEEauthorrefmark{4}mehrday@bgsu.edu,\\
\IEEEauthorrefmark{5}greenr@bgsu.edu,
\IEEEauthorrefmark{6}sniu@bgsu.edu}}
\maketitle

\begin{abstract}
The development of accurate medical image classification models is often constrained by privacy concerns and data scarcity for certain conditions, leading to small and imbalanced datasets. To address these limitations, this study explores the use of generative models, such as Denoising Diffusion Probabilistic Models (DDPM) and Progressive Growing Generative Adversarial Networks (PGGANs), for dataset augmentation.

The research introduces a framework to assess the impact of synthetic images generated by DDPM and PGGANs on the performance of four models: a custom CNN, Untrained VGG16, Pretrained VGG16, and Pretrained ResNet50. Experiments were conducted using Random Sampling and Greedy K Sampling to create small, imbalanced datasets. The synthetic images were evaluated using Frechet Inception Distance (FID) and compared to original datasets through classification metrics.

The results show that DDPM consistently generated more realistic images with lower FID scores and significantly outperformed PGGANs in improving classification metrics across all models and datasets. Incorporating DDPM-generated images into the original datasets increased accuracy by up to 6\%, enhancing model robustness and stability, particularly in imbalanced scenarios. Random Sampling demonstrated superior stability, while Greedy K Sampling offered diversity at the cost of higher FID scores. This study highlights the efficacy of DDPM in augmenting small, imbalanced medical image datasets, improving model performance by balancing the dataset and expanding its size. Our implementation and codes are available at \url{https://github.com/imankhazrak/DDPM_X-Ray}.
\end{abstract}


\maketitle

% Main text
\section{Introduction}

Medical imaging is indispensable in modern healthcare, guiding diagnostics, surgeries, treatment assessments, and disease monitoring. The growing volume of images poses challenges \cite{chan2020deep} for radiologists and physicians to maintain workflow efficiency without technological support. There are significant challenges to train accurate and reliable Machine Learning or Deep Learning diagnostic models. Key issues include the scarcity of extensive and diverse datasets \cite{costa2017towards}, stringent data privacy regulations, and inherent dataset imbalances. These imbalances can lead to biased models that struggle with rare conditions, and even minor errors can have negative implications.

Traditional data augmentation techniques, such as random rotations, flipping, cropping, and noise injection, have been widely employed to expand training datasets. While useful, these methods merely manipulate existing samples and fail to introduce the kind of fundamental variability needed for robust model training \cite{kebaili2023deep}. In contrast, generative models \cite{Frid-Adar.2018_1} , GANs and DDPM, have revolutionized image synthesis by creating entirely new data points. These models offer promising solutions to the challenges of imbalanced datasets, particularly in the field of medical imaging, where the availability of labeled data is limited.

Generative models typically require large and diverse datasets \cite{bandi2023power}. This creates a paradox: if such large labeled datasets were available, effective models could be trained directly. Therefore, generative models are only viable if they can work effectively with scarce datasets. The present paper addresses this challenge by proposing a comprehensive framework for generating synthetic medical images from both small and imbalanced datasets using two generative models: PGGANs \cite{karras2017progressive} and DDPM \cite{ho2020denoising}. We also explore the use of two different sampling methods—Random Sampling and the Greedy K Method—to further assess their impact on model performance. This framework is rigorously evaluated to enhance diagnostic model accuracy and robustness.

\begin{figure*}[ht] 
    \centering
    \includegraphics[width=1\linewidth]{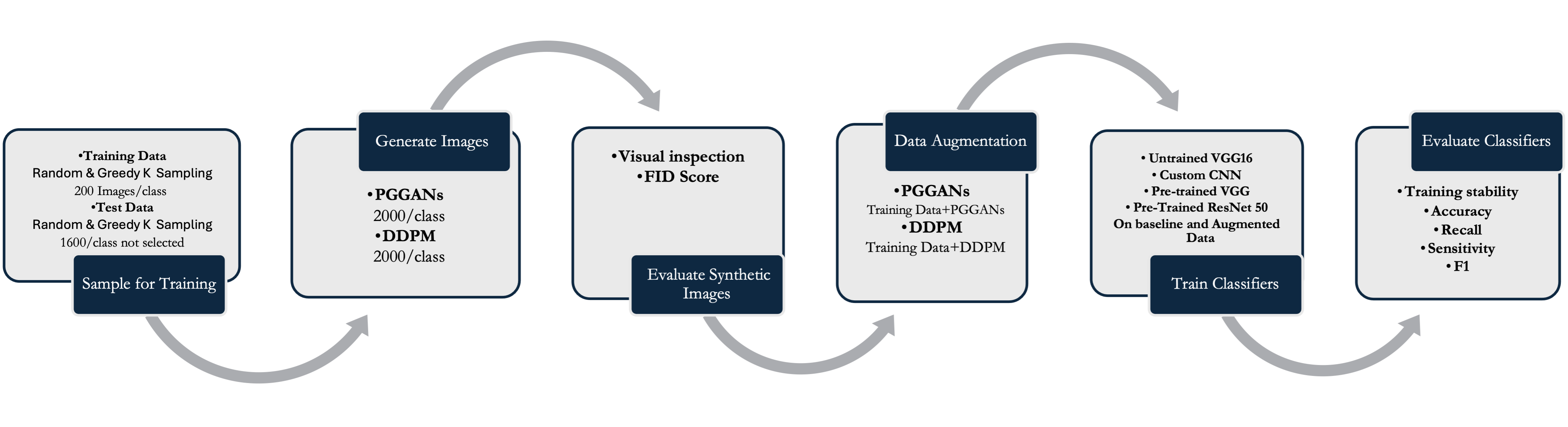}
    \caption{Implemented comprehensive framework that rigorously evaluates the quality and performance of synthetic images generated by DDPM and PGGAN models.} \label{fig: flowchart}
\end{figure*}

This paper introduces a framework for the use of advanced generative models in medical imaging, targeting small and imbalanced datasets. Our key contributions are as follows:

\begin{itemize}
    \item Comprehensive Evaluation Framework: We develop a rigorous framework that evaluates the quality and performance of synthetic images generated by DDPM and PGGANs. This framework consists of three stages. First, synthetic images are generated using both models. Second, the quality of these images is evaluated through visual inspection and quantitative metrics, such as Fréchet Inception Distance (FID), which measures the similarity between generated and real images, and the VGG Expert model for visual validation. Third, we examine the impact of incorporating synthetic images into small and imbalanced datasets on classification model performance.

    \item Generation of High-Quality Synthetic Images: Through extensive experimentation, we demonstrate the feasibility of generating high-quality synthetic images from small medical image datasets using DDPM and PGGANs. DDPM consistently outperforms PGGANs in terms of FID scores, producing more realistic and diverse images that improve dataset size and balance.

    \item Improved Classification Performance: By integrating synthetic images into small and imbalanced datasets, we show significant improvements in the performance of both custom CNNs and untrained VGG16 models. For example, the accuracy of untrained models trained on small datasets improved significantly.

    \item Enhanced Model Stability: Our findings highlight that incorporating synthetic images into the original datasets enhances the stability of both untrained and pretrained classification models. Notably, DDPM provides better stability and consistency in performance than PGGANs, especially under challenging conditions of small dataset.
\end{itemize}

In summary, this study provides a novel approach to overcoming the limitations of small and imbalanced medical datasets by leveraging the latest generative models. It demonstrates the potential of DDPM and PGGANs to augment data in a way that not only improves the size and balance of datasets but also significantly enhances the accuracy, stability, and robustness of classification models in medical imaging. The following sections provide an overview of the generative models, followed by a detailed discussion of DDPM and PGGANs, their applications in medical imaging, and an in-depth analysis of our methodology and results.

The outline of this article consists of  overview of generative models, followed by discussion of two important methods used in the paper, DDPMs and PGGANs, and their applications in medical imaging. Then we proceed to discuss our methodology and obtained results.

%%%%%%%%%%%%%%%%%%%%%%%%%%%%%%%%%%%%%%%%%%
\section{Related Work} \label{RW}

 Generative models, especially those producing high-quality, realistic images, have gained significant attention in augmenting medical datasets, particularly for rare conditions where data scarcity and class imbalance are common. These models can be classified as latent variable generative models, which are either explicit or implicit density models.

 Frameworks from other fields offer valuable insights for healthcare. For example, Sustainability Value Articulation (SVA) enhances internal and external efforts for better social and environmental outcomes by emphasizing supplier involvement and technological integration \cite{jagani2023adopting}. Similarly, the EV supply chain highlights the importance of systematic benchmarking and technological innovation to achieve competitive advantages in complex systems \cite{nejad2024developing}. These principles align with the goals of generative models in addressing data scarcity and improving healthcare datasets, enabling long-term scalability and impact.

Explicit density models like Variational Autoencoders (VAEs), Boltzmann Machines, and DDPMs have predefined density functions, offering interpretability and training stability\cite{kingma2013auto, salakhutdinov2009deep, ho2020denoising}. They are useful for applications like anomaly detection due to their explicit likelihood functions. However, their distributional assumptions can sometimes lead to less realistic images\cite{explicitPros}.

Implicit density models, such as GANs, do not rely on explicit likelihood functions, making them more flexible and capable of modeling complex distributions. They tend to produce more realistic images but suffer from training instability, evaluation difficulty, and sensitivity to hyperparameters \cite{implicitGANS, tang2020GANlessons}.
%=========================================
\subsection{GANs family in Medical Imaging}

GANs are prominent implicit density models that consist of two competing neural networks, a generator, that creates synthetic images from a latent space,  and a discriminator which evaluates resemblance of generated images to real images, engaging in a zero-sum game. Generally, it is hard to train GANs due to training instability \cite {tang2020GANlessons}. PGGANs, introduced by \cite{karras2017progressive}, has significantly improved the stability and quality of GAN-generated images. PGGANs utilize a progressive training approach, starting with low-resolution images and gradually increasing the resolution as training progresses. This technique allows the model to learn coarse features before fine details, leading to more stable training and higher-quality images. 

In medical imaging, GANs mainly have been used to enhance  classification and segmentation deep learning models \cite{kazeminia2020gans}. The work in \cite{costa2017towards} uses GANs on a small CT scan dataset to generate eye fundus images which confirm to the given masks. \cite{mahapatra2018efficient} also used mask to generate lung images, and only those synthetic images that fulfilled informativeness criteria calculated by Bayesian neural networks were used to improve the classifier model. In \cite{Frid-Adar.2018_1}, GANs are employed to synthesize high quality focal liver lesions of multiple conditions to enhance a CNN-classifier.  Moreover, GANs have been successful at synthesizing prostate lesions \cite{kitchen2017deep}, lung cancer nodules \cite{chuquicusma2018fool}, brain MRI images \cite{bermudez2018learning} to name a few. The authors of \cite{baur2018generating} generate high resolution synthetic images of skin lesions from a dataset of 2000 dermoscopic images using multiple GANs architectures and compare their classification performances. They conclude that  PGGANs could to synthesize realistic images that medical professionals upon evaluation were not able to distinguish from real ones. Results of \cite{Beers2018PGGANs} confirm that PGGANs can produce high-resolution images with remarkable detail and consistency, making them one of the best choices for medical image synthesis.

%=========================================
\subsection{Diffusion Family in Medical Imaging}

Diffusion models are generative models that transform noise into structured data through a sequence of steps. The DDPM \cite{ho2020denoising} is a prominent model in this family, known for producing high-fidelity images by reversing a diffusion process. These models iteratively add and then remove noise from an image through two main phases: the forward process, where noise is added over several steps, and the reverse process, where the model learns to denoise the image step-by-step. This iterative refinement allows DDPMs to generate images with fine-grained details. Introduced by Ho et al. in 2020 \cite{ho2020denoising_10}, DDPMs have set new benchmarks in image generation quality by leveraging a sophisticated noise schedule and a robust denoising network.

The application of DDPMs in medical imaging has been explored in various tasks. Nichol and Dhariwal \cite{nichol2021improved} demonstrated the effectiveness of guided diffusion models combined with upsampling techniques in improving MRI resolution, facilitating better diagnosis and treatment planning using a dataset of 10,000 MRI images. In medical imaging, the integration of explainability and trust into AI systems has proven to be crucial for clinician acceptance, particularly for life-threatening diseases like cancer. Rezaeian et al. \cite{rezaeian2024architecture} propose a dual-stage AI model for breast cancer diagnosis, introducing graduated levels of explainability such as tumor localization and probability distributions, which have been shown to significantly enhance trust in AI systems. Similarly, our work focuses on improving the robustness of AI models by addressing data scarcity and imbalance, which are pivotal challenges in building reliable diagnostic tools. Jalal et al. \cite{jalal2021robust} further explored DDPMs for MRI reconstruction, showing significant improvements in image quality and robustness against noise with a dataset of 3,500 MRI images. Similarly, Wolleb et al. \cite{wolleb2022diffusion} applied DDPMs to medical image segmentation, achieving state-of-the-art results with a dataset of 7,500 images. Yoon et al. \cite{Muller2023multimodal} compared latent DDPMs and GANs for medical image synthesis across multiple modalities using 8,000 CT and MRI images, finding DDPMs superior in image quality and diversity. Zhang et al. \cite{mahaulpatha2024ddpm} introduced a DDPM-based X-ray Image Synthesizer using 6,000 X-ray images, demonstrating the model's ability to generate high-fidelity synthetic X-ray images, which can augment training datasets and improve diagnostic model accuracy.

Most studies use large datasets for image generation or do not directly leverage generated datasets to improve model performance. In contrast, our approach uses a small dataset to generate synthetic images and shows how these images enhance model performance, addressing data scarcity and imbalance. This underscores the potential of DDPMs to transform medical imaging, making diagnostic tools more accurate, reliable, and accessible.

%%%%%%%%%%%%%%%%%%%%%%%%%%%%%%%%%%%%%%%%%%
\section{Methodology}
Our research methodology includes several key phases: image synthesis, dataset augmentation, model training and fine-tuning, and performance evaluation.

\subsection{Image Synthesis}

\subsubsection{PGGANs}
\begin{figure}
    \centering
    \includegraphics[width=1\linewidth]{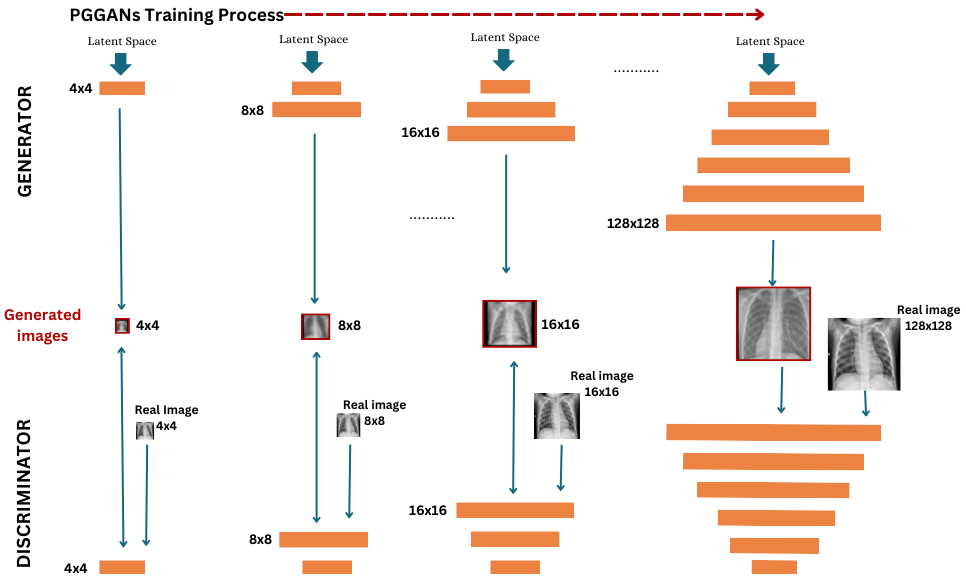}
    \caption{PGGANs gradually increase both the resolution of the generated images and the complexity of the generator and discriminator networks during training.}
    \label{pggan_overview}
\end{figure}
PGGANs utilize a progressive training approach, starting with low-resolution images and gradually increasing the resolution as training progresses (Fig. \ref{pggan_overview}). This method enhances stability and image quality by incrementally increasing the complexity of the generator and discriminator networks. The generator produces data resembling real data, while the discriminator distinguishes between real and generated data \cite{liu2019rob}. The adversarial loss functions for the generator (\( \mathcal{L}_{\text{G}} \)) and discriminator (\( \mathcal{L}_{\text{D}} \)) are:

\[
\mathcal{L}_{\text{G}} = \log(1 - D(G(z)))
\]
\[
\mathcal{L}_{\text{D}} = \log(D(x)) + \log(1 - D(G(z)))
\]

where \( G(z) \) represents the generated data from noise \( z \), and \( D(x) \) represents the discriminator's output for real data \( x \) \cite{alqahtani2021applications}.

PGGANs adopt a step-by-step training approach, beginning with low-resolution images and advancing to higher resolutions. This progressive training allows the model to learn rough features initially and then fine-tune them for generating high-quality images. New layers are added to both networks iteratively, and the loss functions are applied at each resolution level to maintain consistency. 

\subsubsection{DDPM}

\begin{figure}[htb]
    \centering
    \includegraphics[width=1\linewidth]{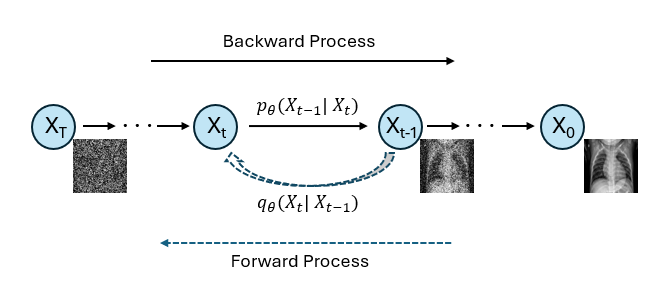}
    \caption{The directed graphical model of DDPM. The forward process gradually adds noise to an image until it becomes pure noise. The backward process then learns to reverse this process, reconstructing the original image from noise.}
    \label{ddpm_overview}
\end{figure}

DDPMs synthesize images by reversing a diffusion process that gradually adds Gaussian noise to an image and then reconstructs the original image from the noise (Fig. \ref{ddpm_overview}) \cite{ho2020denoising}. 

The forward process adds noise to the image:

\[
x_t = \sqrt{\alpha_t} x_{t-1} + \sqrt{1 - \alpha_t} \epsilon_t
\]

where \( x_t \) is the image at iteration \( t \), \( \alpha_t \) is a noise scaling factor, and \( \epsilon_t \) is the Gaussian noise added at iteration \( t \) \cite{ho2020denoising}. The backward process aims to denoise the noisy image obtained from the forward process and recover the original clean image by optimizing the variational lower bound:

\[
L_{DDPM} = \mathbb{E}_{t,x_0,\epsilon}\left[ | \epsilon - \epsilon_{\theta}(x_t, t) |^2 \right]
\]

Here, \( \epsilon \) represents Gaussian noise, and \( \epsilon_{\theta} \) is the noise predicted by the model.

The U-Net architecture, adapted for use in DDPMs, excels in the reverse diffusion process by predicting and removing noise added during the forward phase \cite{ronneberger2015u, baranchuk2021label}. U-Net's U-shaped structure with downsampling and upsampling paths efficiently synthesizes detailed images, incorporating time embeddings to adjust noise prediction based on the reverse process timestep \cite{gong2024pet}.

\subsection{Generated Image Assessment} \label{image_assess}

\paragraph{Visual Inspection}
Generated images are initially evaluated by visually comparing random samples to the original images.

\paragraph{Frechet Inception Distance (FID)}
The FID score quantifies the distributional similarity between real and generated images. It is calculated by extracting features from an InceptionV3 model for both real and generated images, then computing the Frechet distance between the resulting multivariate Gaussian distributions. A higher FID score indicates greater dissimilarity \cite{heusel2017gansFID}.

% \paragraph{\textcolor{red}{VGG16 and ResNet5 Expert Evaluation}}
% \textcolor{red}{To mimic expert radiologist assessment, we employ a VGG Expert classifier model, trained using a VGG16 backbone with pre-trained ImageNet weights on the original data \cite{simonyan2014VGG16}}.

\subsection{Classification Models}
We evaluate the impact of synthetic images using four different classifiers: a pre-trained VGG16, a pre-trained ResNet50, an untrained VGG16, and a custom CNN. Each model is first trained on both the small and imbalanced datasets to establish baselines, followed by training on the same datasets augmented with synthetic images generated by DDPMs and PGGANs. The inclusion of an untrained VGG16 allows us to assess the direct impact of synthetic data on models learning from scratch, providing a clearer understanding of how effective the generated images are in improving generalization without the benefit of pre-learned features. This approach is particularly important in cases where pretrained models may not be available or applicable, and where the goal is to evaluate how synthetic data can help models learn directly from the augmented datasets. To assess stability and observe changes in classification metrics, each model runs five times. This approach allows for a thorough examination of how the augmented datasets influence the models' generalization performance on the test datasets, providing valuable insights into the effectiveness of synthetic data in enhancing both pre-trained and untrained model accuracy.

\begin{table}[htb]
    \centering
    \label{tab:VGG16 architecture}
    \caption{VGG16/ResNet50 Model Summary}
    \resizebox{0.5\textwidth}{!}{
    \begin{tabular}{|l|l|l|}
        \hline
        \textbf{Layer (Type)} & \textbf{Output Shape} & \textbf{Param \#} \\ \hline
        vgg16/resnet50 (\textcolor{blue}{Functional})    & (\textcolor{blue}{None}, \textcolor{green}{7}, \textcolor{green}{7}, \textcolor{green}{512})     & \textcolor{green}{14,714,688}        \\ \hline
        flatten (\textcolor{blue}{Flatten})     & (\textcolor{blue}{None}, \textcolor{green}{25,088})        & \textcolor{green}{0}                 \\ \hline
        dense (\textcolor{blue}{Dense})         & (\textcolor{blue}{None}, \textcolor{green}{512})           & \textcolor{green}{12,845,568}        \\ \hline
        dropout (\textcolor{blue}{Dropout})     & (\textcolor{blue}{None}, \textcolor{green}{512})           & \textcolor{green}{0}                 \\ \hline
        dense\_1 (\textcolor{blue}{Dense})      & (\textcolor{blue}{None}, \textcolor{green}{2})             & \textcolor{green}{1,026}             \\ \hline
        \multicolumn{2}{|l|}{\textbf{Total params:}}      & \textcolor{green}{27,561,282} (105.14 MB)  \\ \hline
        \multicolumn{2}{|l|}{\textbf{Trainable params:}}  & \textcolor{green}{12,846,594} (49.01 MB)   \\ \hline
        \multicolumn{2}{|l|}{\textbf{Non-trainable params:}} & \textcolor{green}{14,714,688} (56.13 MB) \\ \hline
    \end{tabular}
    }
    
\end{table}

\begin{table}[htb]
    \centering
    \caption{Custom CNN Model Summary}
    \label{tab:Model architecture}
    \resizebox{0.5\textwidth}{!}{
    \begin{tabular}{|l|l|l|}
        \hline
        \textbf{Layer (Type)} & \textbf{Output Shape} & \textbf{Param \#} \\ \hline
        conv2d\_1 (\textcolor{blue}{Conv2D})    & (\textcolor{blue}{None}, \textcolor{green}{128}, \textcolor{green}{128}, \textcolor{green}{64})   & \textcolor{green}{1,792}         \\ \hline
        batch\_normalization\_1 (\textcolor{blue}{BatchNormalization}) & (\textcolor{blue}{None}, \textcolor{green}{128}, \textcolor{green}{128}, \textcolor{green}{64})  & \textcolor{green}{256}           \\ \hline
        max\_pooling2d\_1 (\textcolor{blue}{MaxPooling2D})   & (\textcolor{blue}{None}, \textcolor{green}{64}, \textcolor{green}{64}, \textcolor{green}{64})    & \textcolor{green}{0}             \\ \hline
        conv2d\_2 (\textcolor{blue}{Conv2D})  & (\textcolor{blue}{None}, \textcolor{green}{64}, \textcolor{green}{64}, \textcolor{green}{128})   & \textcolor{green}{73,856}        \\ \hline
        batch\_normalization\_2 (\textcolor{blue}{BatchNormalization}) & (\textcolor{blue}{None}, \textcolor{green}{64}, \textcolor{green}{64}, \textcolor{green}{128})   & \textcolor{green}{512}           \\ \hline
        max\_pooling2d\_2 (\textcolor{blue}{MaxPooling2D}) & (\textcolor{blue}{None}, \textcolor{green}{32}, \textcolor{green}{32}, \textcolor{green}{128}) & \textcolor{green}{0}             \\ \hline
        conv2d\_3 (\textcolor{blue}{Conv2D})  & (\textcolor{blue}{None}, \textcolor{green}{32}, \textcolor{green}{32}, \textcolor{green}{256})   & \textcolor{green}{295,168}       \\ \hline
        batch\_normalization\_3 (\textcolor{blue}{BatchNormalization}) & (\textcolor{blue}{None}, \textcolor{green}{32}, \textcolor{green}{32}, \textcolor{green}{256})   & \textcolor{green}{1,024}         \\ \hline
        max\_pooling2d\_3 (\textcolor{blue}{MaxPooling2D}) & (\textcolor{blue}{None}, \textcolor{green}{16}, \textcolor{green}{16}, \textcolor{green}{256}) & \textcolor{green}{0}             \\ \hline
        flatten (\textcolor{blue}{Flatten})   & (\textcolor{blue}{None}, \textcolor{green}{65536})        & \textcolor{green}{0}             \\ \hline
        dense\_1 (\textcolor{blue}{Dense})       & (\textcolor{blue}{None}, \textcolor{green}{256})          & \textcolor{green}{16,777,472}    \\ \hline
        dropout\_1 (\textcolor{blue}{Dropout})   & (\textcolor{blue}{None}, \textcolor{green}{256})          & \textcolor{green}{0}             \\ \hline
        dense\_2 (\textcolor{blue}{Dense})    & (\textcolor{blue}{None}, \textcolor{green}{128})          & \textcolor{green}{32,896}        \\ \hline
        dropout\_2 (\textcolor{blue}{Dropout})& (\textcolor{blue}{None}, \textcolor{green}{128})          & \textcolor{green}{0}             \\ \hline
        dense\_3 (\textcolor{blue}{Dense})    & (\textcolor{blue}{None}, \textcolor{green}{2})            & \textcolor{green}{258}           \\ \hline
        \multicolumn{2}{|l|}{\textbf{Total params:}}        & \textcolor{green}{17,183,234} (65.55 MB)   \\ \hline
        \multicolumn{2}{|l|}{\textbf{Trainable params:}}    & \textcolor{green}{17,182,338} (65.55 MB)   \\ \hline
        \multicolumn{2}{|l|}{\textbf{Non-trainable params:}} & \textcolor{green}{896} (3.50 KB)          \\ \hline
    \end{tabular}
    }
    
\end{table}

%%%%%%%%%%%%%%%%%%%%%%%%%%%%%%%%%%%%%%%%%%
\section{Experimental Results}

This section presents the findings from our experiments. 

\paragraph{Computation Resources}
Our experiments are conducted on a Pitzer GPU cluster node from OSC (Ohio Supercomputer Center) with Dual NVIDIA Volta V100 GPUs with 32 GB GPU memory and 48 cores per node at 2.9 GHz. We used Python for the implementation, PyTorch for the generative models and TensorFlow for the classification models.

\paragraph{Dataset}
The original dataset for this study, sourced from Kaggle, consists of Chest X-ray (CXR) images categorized into two classes: 1,802 NORMAL and 1,800 PNEUMONIA. Each image is originally 256x256 pixels. However, in order to simulate real-world scenarios, two types of datasets are created: small and imbalanced datasets.

\begin{itemize}
    \item \textbf{Small Dataset:} We sample 200 images from each class (NORMAL and PNEUMONIA) to create a balanced training dataset. The remaining 1,600 images per class form the test set. This small dataset helps in evaluating model performance under conditions where data availability is limited, mimicking clinical situations with rare medical conditions.

    \item \textbf{Imbalanced Dataset:} We construct an imbalanced dataset by sampling 1,500 images from the NORMAL class and 200 images from the PNEUMONIA class for training. For testing, we create three separate imbalanced test sets by randomly sampling 300 images from the NORMAL class and 100 images from the PNEUMONIA class. The models are tested on all three test sets, and the average performance metrics are computed to ensure reliable comparisons. This imbalanced dataset reflects the distribution often found in medical datasets, where certain conditions (like pneumonia) are underrepresented.

\end{itemize}

To ensure diversity and robustness, two different sampling methods are employed:
\begin{enumerate}
    \item \textbf{Random Sampling:} Images are selected randomly from the larger dataset, closely mimicking real-world datasets where the available data is typically uncurated and randomly distributed. This method ensures a more natural distribution of images but may not always capture the diversity of the dataset, which can limit the performance of generative models.

    \item \textbf{Greedy K Sampling:} Images are selected based on their dissimilarity to others, ensuring a more diverse representation within the training set. This method helps to reduce computational cost by focusing on a smaller, highly diverse subset of the data, making it more efficient to generate synthetic images. The increased diversity of the selected data leads to the generation of synthetic images that exhibit greater variability, ultimately improving model generalization.
\end{enumerate}

The combination of these approaches results in four distinct datasets: a small and an imbalanced dataset for each sampling method. These datasets are then used for training the classification models and assessing the impact of synthetic images generated by DDPM and PGGANs.(TABLE \ref{tab:datasets})

\begin{table}[htb]
\centering
\caption{Dataset Overview}
\resizebox{0.5\textwidth}{!}{
\begin{tabular}{|l|c|c|c|}
\hline
\multirow{2}{*}{\textbf{Dataset Type}}          &\multirow{2}{*} {\textbf{Sampling Method}} & \textbf{Training Data} & \textbf{Test Data} \\ 
                               &                          & (NORMAL, PNEUMONIA) & (NORMAL, PNEUMONIA) \\ \hline
Original Dataset               & -                        & 1802, 1800             & -                             \\ \hline
\multirow{2}{*}{Small Dataset} & Random                   & 200, 200               & 1602, 1600                     \\ \cline{2-4} 
                               & Greedy K                 & 200, 200               & 1602, 1600                     \\ \hline
\multirow{4}{*}{Imbalanced Dataset} & \multirow{2}{*}{Random}                    & \multirow{2}{*}{1500, 300}              & 3 Imbalanced Test Sets        \\ 
                               &                          &                        & (300, 100) each                   \\ \cline{2-4} 
                               &  \multirow{2}{*}{Greedy K}                 & \multirow{2}{*}{1500, 300}              & 3 Imbalanced Test Sets        \\ 
                               &                          &                        & (300, 100) each                   \\ \hline
\end{tabular}
\label{tab:datasets}
}
\end{table}

\subsection{Synthetic Image Generation} 
The PGGANs and DDPM models are trained separately for each class in the training dataset, producing a total of four models using 200 images from the small dataset for each sampling method. Leveraging the code from \cite{ProgressiveGANPyTorch} and \cite{UNet2D_Diffusers}, we generate 2,000 images per class for each model.

\begin{figure}[htb]
    \centering
    \includegraphics[width=1\linewidth]{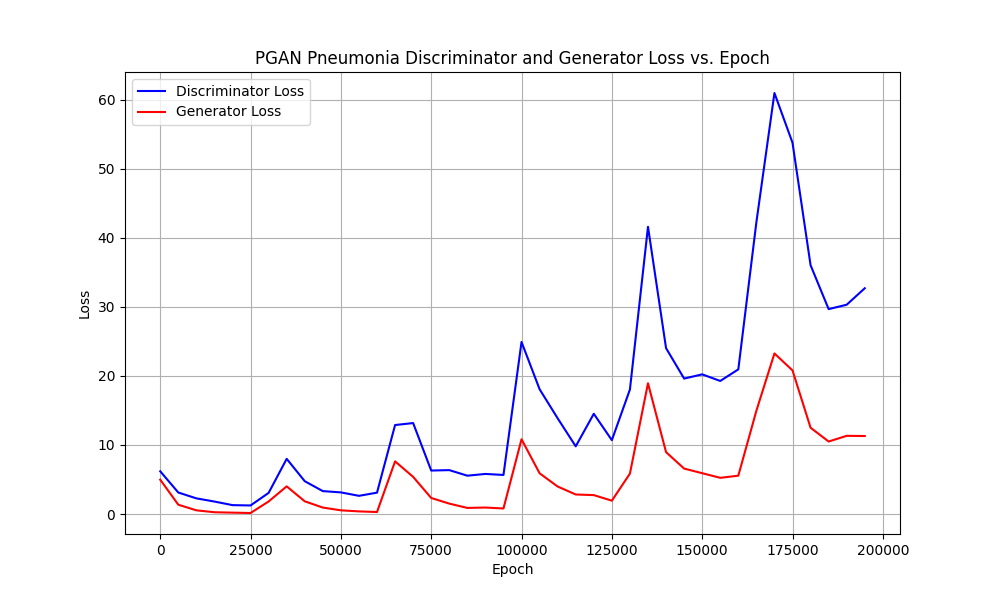}
    \caption{PGGANs model training loss for Pneumonia class. Plot indicates the training process is stable where loss function increases as model adds new layers.}.
    \label{fig: PGGANsLoss}
\end{figure}

To train PGGANs models, random sampling from a standard normal distribution is employed for initialization. Stability in training is achieved by equalizing the learning rate, i.e. scaling the outputs right before the forward pass \cite{karras2017progressive}. Convolution layers below 64-pixel resolution are set at 128 filters, while layers at 64 and 128-pixel resolution are set to 64 filters. The BATCH-SIZE is set to 4. One PGGANs model is trained per class using the Adam optimizer and the Wasserstein loss, each for 200,000 epochs. Due to computational constraints, the models did not converge, though the training process was stable and followed desired pattern of loss (Fig \ref{fig: PGGANsLoss}). With experimental trials relying on computed losses, we choose the checkpoint from epochs 160000 (PGGANs 160k).

The DDPM model hyperparameters include an image size of 128 pixels, a batch size of 16, a learning rate of 1e-4, 512 epochs, 8000 timesteps, and mixed precision ("fp16") to reduce memory use and speed up data transfer.

% We generate 2000 images per class from each model. It is important to note that the computational cost for generating synthetic images with DDPM model after training is high. In contrast, for PGGANs, once the model is trained, image generation is almost instantaneous.  

\subsection{Synthetic Image Quality Evaluation}
\subsubsection{Visual Inspection}
\begin{figure}[htb]
    \centering
    \includegraphics[width=1\linewidth]{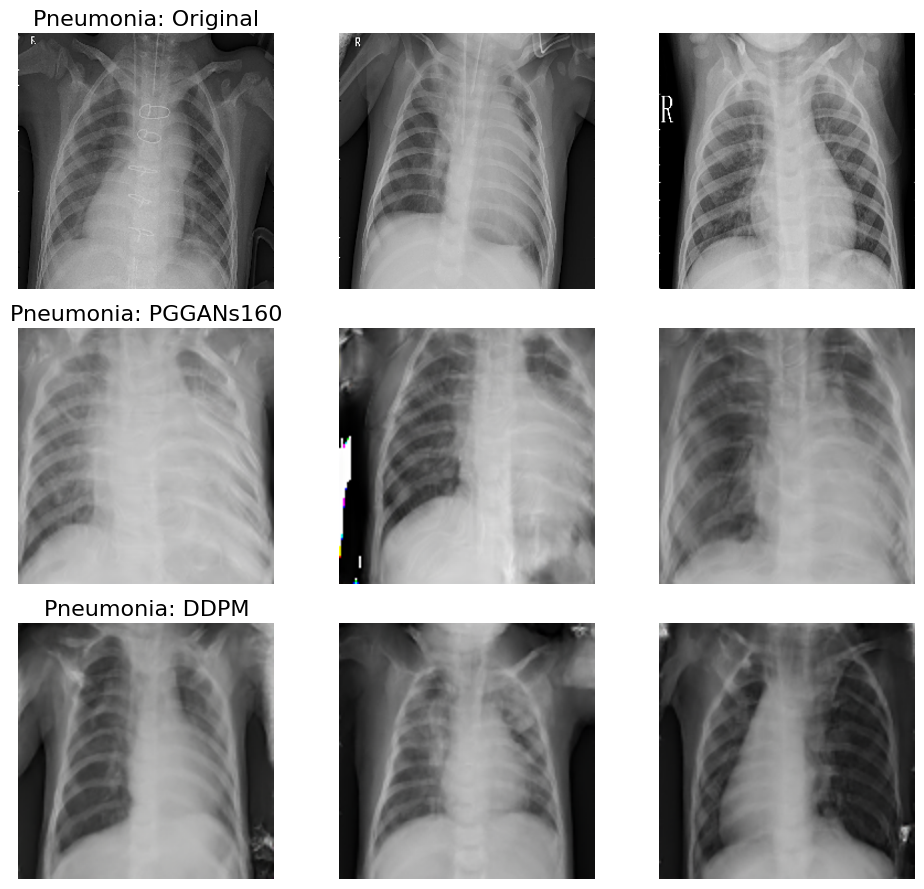}
    \caption{Comparison between Original vs. DDPM  vs. PGGANs 160K Pneumonia class images. Synthetic images exhibit higher similarity to the originals.}
    \label{fig: Pneumonia150k}
\end{figure}

\begin{figure}[htb]
    \centering
    \includegraphics[width=1\linewidth]{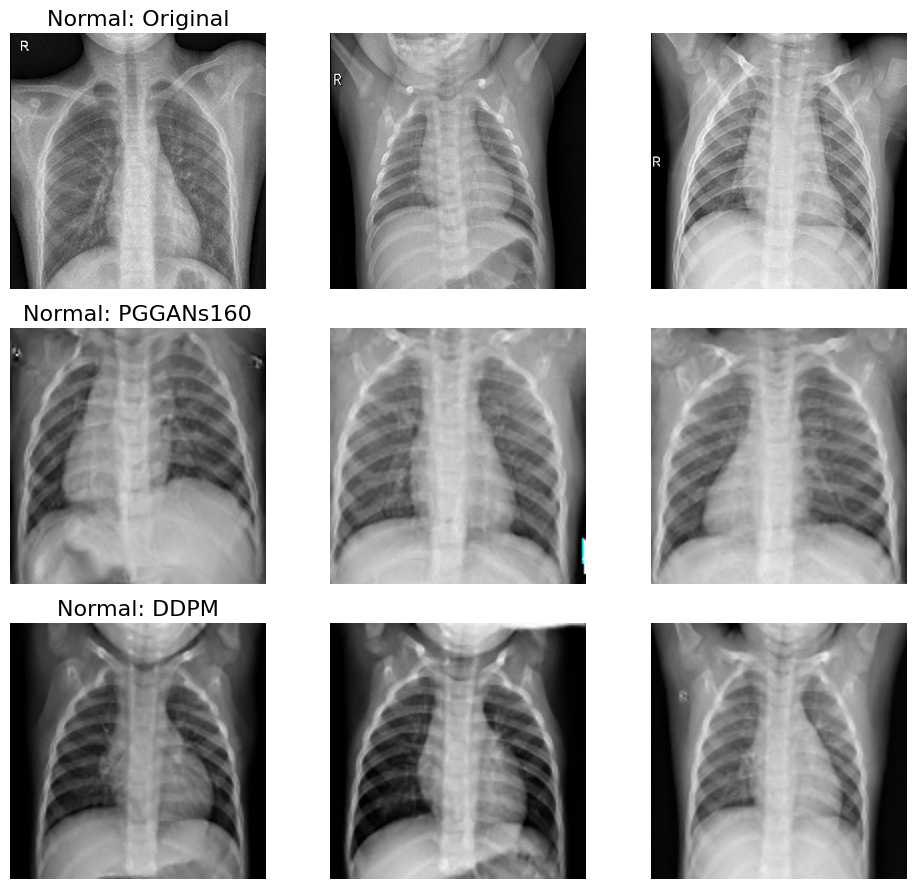}
    \caption{Comparison between Original vs. DDPM  vs. PGGANs 160K NORMAL class images. Synthetic images exhibit higher similarity to the originals.}
    \label{fig: Normal160k}
\end{figure}

Figs \ref{fig: Pneumonia150k} and \ref{fig: Normal160k} showcase a visual comparison of generated images of both healthy and pneumonia-affected lungs. Although visual inspection can be subjective, the PGGANs images from the 160k checkpoint are visually appealing but occasionally display defect patches. In contrast, the DDPM-generated images demonstrate a closer resemblance to the original data, exhibiting superior visual fidelity.

%-----------------------------------------

\subsubsection{FID Metric} 
\begin{figure}[htb]
    \centering
    \includegraphics[width=1\linewidth]{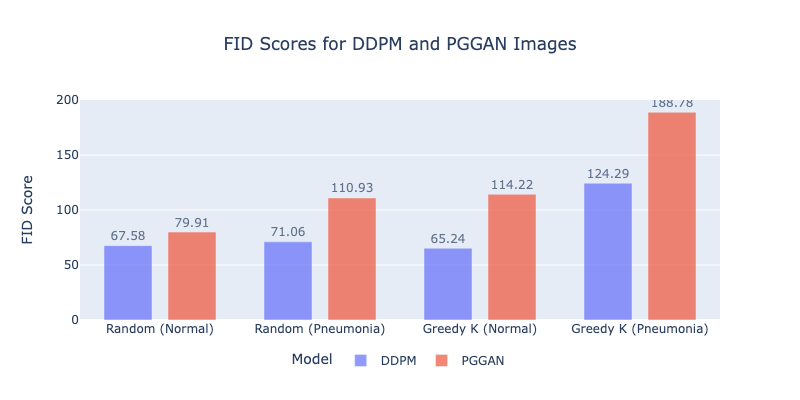}
    \caption{Comparison of FID Scores between original and synthetically generated images using DDPM and PGGANs.}
    \label{fig: fid_metric}
\end{figure}

PyTorch implementation provided by \cite{Seitzer2020FID} is used to calculate FID scores between original data set and each model's generated images per class (Fig.\ref{fig: fid_metric}).

Fig.\ref{fig: fid_metric} shows FID metrics for the DDPM and PGGANs models in two scenarios—Random and Greedy K sampling methods—after generating 2000 synthetic images for both the NORMAL and PNEUMONIA labels.

Across both sampling methods, the FID scores for DDPM are consistently lower than those for PGGANs, indicating that DDPM generates more realistic synthetic images that are closer to the real data distribution. In both models, the FID values for the PNEUMONIA label are higher than those for the NORMAL label, suggesting that generating realistic PNEUMONIA images is more challenging.

However, the impact of the sampling method is evident in the significantly higher FID scores observed with the Greedy K method compared to Random Sampling, especially for PGGANs. The Greedy K method selects more diverse and distinct samples, which encourages the generative models to produce a wider variety of images, including rare or uncommon patterns in the dataset. While this increases image diversity, it also makes it harder for the models to maintain fidelity to the real data, leading to higher FID scores—particularly for the PNEUMONIA label, which is more difficult to generate accurately.

% In summary, DDPM outperforms PGGANs across both sampling methods, and the increased difficulty of generating realistic images under Greedy K sampling reflects the trade-off between diversity and fidelity in synthetic image generation.

Here’s a more professional version of your section:

\subsection{Classification Models}

To assess the impact of data augmentation using generative models, we conducted two distinct experiments using two different sampling methods: Random Sampling and the Greedy K Method. For each sampling method, we performed two experiments—one on a small dataset and another on an imbalanced dataset.

Four models were evaluated in each experiment: a custom CNN (trained for 20 epochs), an untrained VGG16 (10 epochs), a pretrained VGG16 (5 epochs), and a pretrained ResNet50 (5 epochs). The pretrained weights for both the VGG16 and ResNet50 models were sourced from the ImageNet Large Scale Visual Recognition Challenge (ILSVRC) dataset, which consists of 1.2 million images categorized into 1,000 classes. The following datasets were used for model training for each sampling method:

\begin{enumerate} 
    \item Small Dataset 
    \begin{itemize} 
        \item \textbf{Baseline}: Training dataset with a total size of 400 images. 
        \item \textbf{DDPM Mix}: Training dataset augmented with DDPM-generated synthetic images, resulting in a total size of 4,400 images. 
        \item \textbf{PGGANs Mix}: Training dataset augmented with PGGANs-generated synthetic images, resulting in a total size of 4,400 images. 
    \end{itemize} 
    \item Imbalanced Dataset 
    \begin{itemize} 
        \item \textbf{Baseline}: Training dataset with a total of 1,700 images (1,500 NORMAL, 200 PNEUMONIA). 
        \item \textbf{DDPM Mix}: Training dataset augmented with DDPM-generated synthetic images, leading to 3,700 images (1,500 NORMAL, 2,200 PNEUMONIA). 
        \item \textbf{PGGANs Mix}: Training dataset augmented with PGGANs-generated synthetic images, leading to 3,700 images (1,500 NORMAL, 2,200 PNEUMONIA). 
    \end{itemize} 
\end{enumerate}

All hyperparameters were kept consistent across experiments. Models were evaluated using accuracy, recall, precision, and F1 score. To ensure robustness and model stability, each model was run 5 times, with training and validation data shuffled for each run.

\vspace{0.3cm}
\subsubsection{Random Sampling}
We employed random sampling to generate both small and imbalanced datasets, simulating real-world scenarios where data distribution is often uneven and limited in size. In this section, we present the results of these experiments, highlighting the impact of synthetic data augmentation on model performance under these challenging conditions (TABLES \ref{tab: accuracy random balanced data}, \ref{tab: accuracy random imbalanced data} and Fig \ref{fig: Classification}).

\begin{table}[htb]
    \centering
    \caption{Accuracy with Standard Deviation (SD) for Models Using Random Method on Balanced Dataset}
    \label{tab: accuracy random balanced data}
    \resizebox{0.5\textwidth}{!}{
    \begin{tabular}{|c|c|c|c|}
        \hline
        \textbf{Model}         & \textbf{Original} & \textbf{DDPM} & \textbf{PGGANs} \\ \hline
        Custom CNN     & 0.90 $\pm$ 0.035  & \textbf{0.93 $\pm$ 0.011}  & 0.92 $\pm$ 0.011     \\ \hline
        Untrained VGG16& 0.86 $\pm$ 0.039  & \textbf{0.92 $\pm$ 0.018}  & 0.88 $\pm$ 0.016     \\ \hline
        Pretrained VGG16& 0.93 $\pm$ 0.007 & \textbf{0.95 $\pm$ 0.005}  & 0.92 $\pm$ 0.019     \\ \hline
        Pretrained ResNet50 & 0.93 $\pm$ 0.015 & 0.93 $\pm$ 0.017  & \textbf{0.94 $\pm$ 0.002}     \\ \hline
    \end{tabular}
    }
\end{table}

\begin{table}[htb]
\centering
\caption{Accuracy with Standard Deviation (SD) for Models Using Random Method on Imbalanced Dataset}
\label{tab: accuracy random imbalanced data}
\resizebox{0.5\textwidth}{!}{
\begin{tabular}{|c|c|c|c|}
\hline
\textbf{Model}         & \textbf{Original} & \textbf{DDPM} & \textbf{PGGANs} \\ \hline
Custom CNN     & 0.95 $\pm$ 0.014  & \textbf{0.95 $\pm$ 0.011}  & 0.95 $\pm$ 0.018     \\ \hline
Untrained VGG16& 0.86 $\pm$ 0.028  & \textbf{0.92 $\pm$ 0.008}  & 0.89 $\pm$ 0.030     \\ \hline
Pretrained VGG16& 0.94 $\pm$ 0.019 & 0.94 $\pm$ 0.019  & \textbf{0.94 $\pm$ 0.018}     \\ \hline
Pretrained ResNet50 & 0.95 $\pm$ 0.017 & \textbf{0.95 $\pm$ 0.010}  & 0.95 $\pm$ 0.016     \\ \hline
\end{tabular}
}
\end{table}

\textbf{Small Dataset:}

    \begin{itemize}
        \item The results for the Custom CNN model show a clear improvement when using both DDPM and PGGANs-augmented datasets compared to the original dataset. On the original dataset, the model achieved an average accuracy, F1 score, recall, and precision of 0.90, with a standard deviation of 0.035, indicating moderate variability between runs. With the DDPM-augmented dataset, these metrics improved to 0.93, and the standard deviation dropped significantly to 0.011, reflecting more consistent performance. The PGGANs-augmented dataset also enhanced performance, with accuracy and other metrics increasing to 0.92. The variability was similarly reduced, with a standard deviation of 0.011.
        
        \item The results for the Untrained VGG16 model show significant improvements when using both DDPM and PGGANs-augmented datasets compared to the original dataset. With the DDPM-augmented dataset, the average accuracy, F1 score, recall, and precision improved substantially to 0.92. The standard deviation decreased to 0.021, indicating more consistent results across runs. This reflects a significant enhancement in both performance and stability due to the DDPM-generated synthetic images.

        \item The Pretrained VGG16 model performed strongly across all datasets, with the highest accuracy and stability observed with the DDPM-augmented dataset. The average accuracy increased to 0.95. The standard deviation remained low at 0.017, showing highly consistent performance.

        \item The Pretrained ResNet50 model performed best on the PGGANs-augmented dataset, achieving an average accuracy of 0.94. The standard deviation was 0.002, indicating high stable performance across multiple runs.
    \end{itemize}

\textbf{Imbalanced Dataset}:  

    \begin{itemize}
        \item While the accuracy for Custom CNN remained consistent across all datasets, the DDPM augmentation provided better stability with lower standard deviation compared to the original and PGGANs-augmented datasets. PGGANs increased variability slightly, making DDPM the better option for stability.

        \item For the untrained VGG16 model with the DDPM-augmented dataset, accuracy saw a substantial improvement, reaching 0.92. Additionally, the standard deviation dropped to 0.009, indicating more consistent performance across runs and demonstrating enhanced stability compared to both the original dataset and the PGGAN-augmented dataset.

        \item For both the pretrained VGG16 and ResNet50 models, the augmented datasets showed a reduction in standard deviation, indicating enhanced stability and more consistent performance across runs.
    \end{itemize}

\vspace{0.3cm}

\subsubsection{Greedy K Sampling}  
  
\begin{figure*}[htb]
    \centering
    \includegraphics[width=1\linewidth]{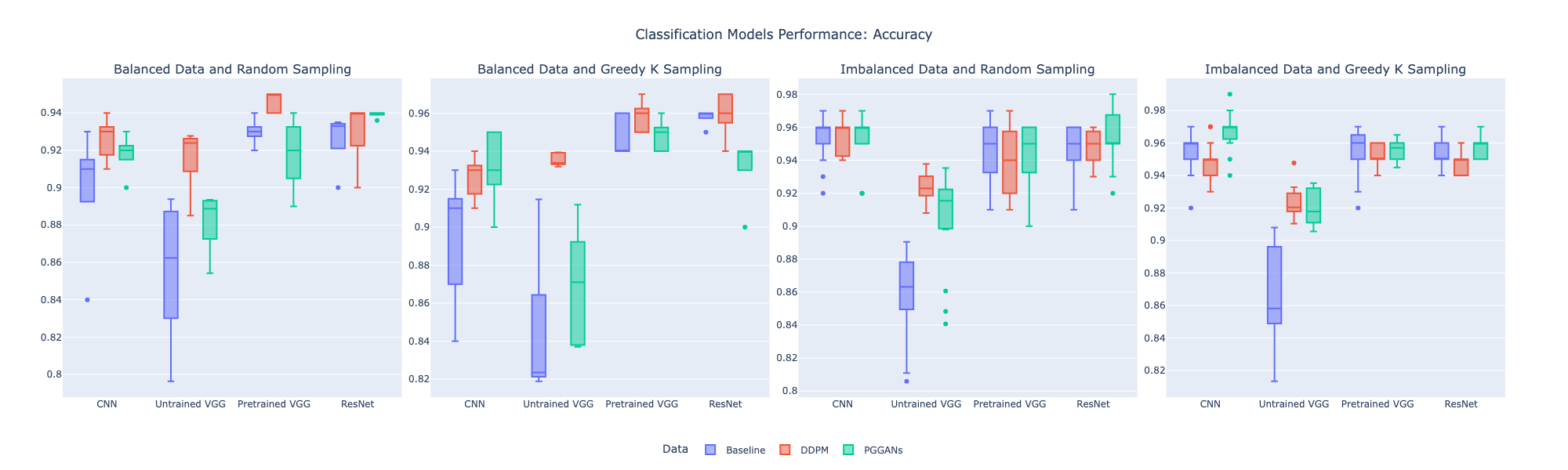}
    \includegraphics[width=1\linewidth]{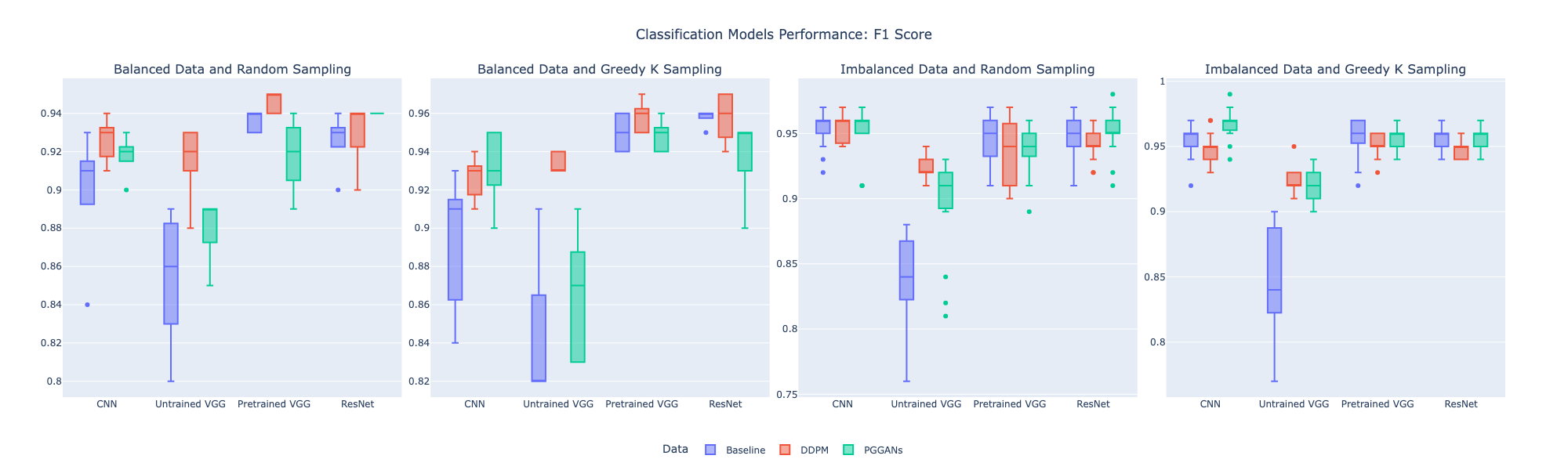}
    \caption{Accuracy and F1 Scores for classification models on different datasets.}
    \label{fig: Classification}
\end{figure*}

In the next experiment, we used the Greedy K method to create small and imbalanced datasets and this section highlights the impact of synthetic data augmentation on model performance (TABLES \ref{tab: accuracy greedy balanced data}, \ref{tab: accuracy greedy imbalanced data} and Fig \ref{fig: Classification}).

\begin{table}[htb]
\centering
\caption{Accuracy with Standard Deviation (SD) for Models Using Greedy K Method on Balanced Dataset}
\resizebox{0.5\textwidth}{!}{
\label{tab: accuracy greedy balanced data}
\begin{tabular}{|c|c|c|c|}
\hline
\textbf{Model}         & \textbf{Original} & \textbf{DDPM} & \textbf{PGGANs} \\ \hline
Custom CNN     & 0.89 $\pm$ 0.035  & \textbf{0.93 $\pm$ 0.011}  & 0.93 $\pm$ 0.020     \\ \hline
Untrained VGG16& 0.85 $\pm$ 0.040  & \textbf{0.94 $\pm$ 0.003 } & 0.87 $\pm$ 0.032     \\ \hline
Pretrained VGG16& 0.95 $\pm$ 0.011 & \textbf{0.96 $\pm$ 0.008}  & 0.95 $\pm$ 0.008     \\ \hline
Pretrained ResNet50 & \textbf{0.96 $\pm$ 0.004} & 0.96 $\pm$ 0.012  & 0.93 $\pm$ 0.018     \\ \hline
\end{tabular}
}
\end{table}

\begin{table}[htb]
    \centering
    \caption{Accuracy Results for Models Using Greedy K Method on Imbalanced Dataset with Standard Deviation}
    \label{tab: accuracy greedy imbalanced data}
    \resizebox{0.5\textwidth}{!}{
        \begin{tabular}{|c|c|c|c|}
        \hline
        \textbf{Model}         & \textbf{Original} & \textbf{DDPM} & \textbf{PGGANs} \\ \hline
        Custom CNN     & 0.95 $\pm$ 0.014  & 0.95 $\pm$ 0.012  & \textbf{0.97 $\pm$ 0.012}     \\ \hline
        Untrained VGG16& 0.87 $\pm$ 0.028  & \textbf{0.92 $\pm$ 0.010}  & 0.92 $\pm$ 0.011     \\ \hline
        Pretrained VGG16& 0.96 $\pm$ 0.016 & \textbf{0.97 $\pm$ 0.007}  & 0.96 $\pm$ 0.008     \\ \hline
        Pretrained ResNet50 & \textbf{0.95 $\pm$ 0.006} & \textbf{0.96 $\pm$ 0.006}  & 0.96 $\pm$ 0.007     \\ \hline
        \end{tabular}
    }
\end{table}

\begin{itemize}
    \item The Custom CNN model saw a significant boost in both accuracy and stability with the augmented datasets. Accuracy increased from 0.89 (original) to 0.93 (DDPM), with the standard deviation dropping from 0.035 to 0.011, indicating more stable performance.
     
    \item For the untrained VGG16, DDPM significantly boosted accuracy from 0.85 (original) to 0.94, with the standard deviation decreasing from 0.040 to 0.003, demonstrating much greater stability with DDPM augmentation.
    
    \item The pretrained VGG16 model saw an accuracy increase from 0.95 to 0.96 with the DDPM-augmented dataset, accompanied by a reduction in standard deviation from 0.011 to 0.008, indicating improved stability.

    \item For the pretrained ResNet50, the DDPM dataset maintained the same high accuracy of 0.96 as the original dataset, with no improvement in model stability observed after adding synthetic images.
\end{itemize}

\textbf{Imbalanced Dataset}: 

\begin{itemize}
    \item For Custom CNN, both DDPM and PGGANs improved accuracy, with PGGANs reaching the highest accuracy of 0.97, but DDPM offered better stability.
    \item For Untrained VGG16, both augmentations significantly improved accuracy to 0.92, with DDPM providing better stability due to lower variability.
    \item For the pretrained VGG16, PGGANs resulted in a slight accuracy increase to 0.96, while the standard deviation dropped to 0.005, indicating enhanced stability compared to the original dataset.
    \item Finally, for the pretrained ResNet50, PGGANs led to a slight improvement in both accuracy and stability.
\end{itemize}

The improvement in robustness by adding synthetic images can be attributed to the increased diversity and balance introduced to the training data. In small and imbalanced datasets, models are often prone to overfitting, learning patterns that are not generalizable due to the lack of variety in the data. By incorporating synthetic images, especially those generated by DDPM, the training dataset is expanded with new, diverse samples that better represent the underlying data distribution. This reduces overfitting, enabling the model to learn more generalized features. Additionally, the added synthetic data helps balance the dataset, ensuring that minority classes are adequately represented, which is crucial for improving the model's ability to correctly classify rare conditions. Together, these factors contribute to the enhanced stability and accuracy observed across multiple runs, making the models more robust to variations in the data.

 %%%%%%%%%%%%%%%%%%%%%%%%%%%%%%%%%%%%%%%%%%
\section{Conclusion}

The experiments conducted in this study evaluated the effects of data augmentation using DDPM and PGGANs on small and imbalanced medical image datasets, utilizing two sampling methods: Random Sampling and Greedy K. The use of generative models for data augmentation led to significant improvements in both the accuracy and robustness of the models. Across all experiments, the introduction of synthetic images enhanced performance, with models exhibiting greater stability and reduced variability during training. This resulted in higher accuracy and better generalization, particularly when handling limited and imbalanced datasets. Of the two generative models, DDPM consistently outperformed PGGANs, delivering superior gains in accuracy and robustness. DDPM-generated images more closely aligned with the original data distribution, contributing to more stable performance and lower variability across all datasets and sampling methods..

The FID evaluation showed that DDPM outperformed PGGANs with lower FID scores, indicating that DDPM-generated images more closely resembled the original data. Random Sampling led to lower FID scores across both models, suggesting better fidelity, while the Greedy K method, though encouraging diversity, resulted in higher FID scores, particularly for PGGANs.

In terms of performance on small and imbalanced datasets, DDPM consistently improved accuracy and reduced variability across models. For instance, the Custom CNN model's accuracy increased to 0.93 on the small dataset and 0.95 on the imbalanced dataset with DDPM, both showing reduced variability. In contrast, while PGGANs increased accuracy, it also introduced higher variability, making it less stable. This was particularly evident for the Untrained VGG16 model, where DDPM raised accuracy to 0.94 with minimal variability, while PGGANs struggled to reduce variability.

Random Sampling proved to be the most stable approach, providing consistent improvements in accuracy and reduced variability across models when paired with DDPM-generated images. This method reflects real-world scenarios, where data is often limited and randomly distributed, making it a practical choice when working with small datasets. On the other hand, Greedy K Sampling—which selects a subset of diverse images—offers a different advantage. While it led to higher variability (higher FID), especially for PGGANs, it still achieved high accuracy when combined with DDPM, as seen in the Pretrained VGG16, which reached 0.97 accuracy with minimal variability. Greedy K is particularly beneficial when we aim to train generative models on a smaller, more diverse subset of images, reducing computational costs while generating synthetic images that not only promote diversity but also help the model achieve greater stability and accuracy. This method allows for efficient use of resources while ensuring that the generated data adds meaningful variety to the dataset, ultimately improving model performance.

The improvement in robustness from adding synthetic images comes from the increased diversity and balance they introduce. In small and imbalanced datasets, models are prone to overfitting due to limited variety. Synthetic images, especially from DDPM, expand the dataset with diverse samples, reducing overfitting and helping the model generalize better. The balanced data also ensures better representation of minority classes, improving classification of rare conditions. These factors together enhance stability, accuracy, and robustness across multiple runs.

In summary, generative models like DDPM and PGGANs offer an effective solution for handling the challenges of small and imbalanced datasets by generating synthetic images that enhance model performance. These models help improve accuracy, increase data diversity, and reduce variability, leading to more robust and stable training. While Greedy K Sampling encouraged diversity, Random Sampling proved more effective in maintaining stability, making generative models a reliable option for augmenting medical image datasets.

This study highlights the efficacy of generative models like DDPM and PGGANs in addressing challenges associated with small and imbalanced medical datasets. Future research could expand on this by integrating interdisciplinary strategies to enhance healthcare outcomes. For instance, incorporating clinician stress indicators could improve both diagnostic accuracy and well-being \cite{wang2024relationship}. Furthermore, innovation through Corporate Social Responsibility (CSR) emphasizes fostering trust and improving collaborative processes, which could inform strategies for integrating generative models in healthcare \cite{callaway2023innovating}. Advancements in IT demonstrate the potential of applying diversification strategies to optimize system performance, while theoretical frameworks for scalable technology adoption, as outlined in \cite{callaway2024drives}, provide a foundation for integrating generative methods effectively. Additionally, methods like diversity indices, such as the Shannon Diversity Index, could be used to assess the impact of generative models on healthcare datasets, helping to balance specialized data generation with broader themes in medical imaging \cite{mashhadi2024refining}. Together, these approaches could create a more comprehensive and impactful application of generative models in medical imaging.

\bibliographystyle{IEEEtran}
% \bibliography{references}

\end{document}